# Threat Detection In Self Driving Vehicles Using Computer Vision

Umang Goenka[1], Aaryan Jagetia[1] Param Patil[2], Akshay Singh[3], Taresh Sharma[3], and Poonam Saini[3]

1. Indian Institute of Information Technology, Lucknow, India.
   1. lit2019033@iiitl.ac.in,
2. Sardar Patel Institute Of Technology, Mumbai, India.
3. Punjab Engineering College, Chandigarh, India

**Abstract.** On-road obstacle detection is an important field of research that falls in the scope of intelligent transportation infrastructure systems. The use of vision-based approaches results in an accurate and costeffective solution to such systems. In this research paper, we propose a threat detection mechanism for autonomous self-driving cars using dashcam videos to ensure the presence of any unwanted obstacle on the road that falls within its visual range. This information can assist the vehicle's program to en route safely. There are four major components, namely, YOLO to identify the objects, advanced lane detection algorithm, multi regression model to measure the distance of the object from the camera, the two-second rule for measuring the safety, and limiting speed. In addition, we have used the Car Crash Dataset(CCD) for calculating the accuracy of the model. The YOLO algorithm gives an accuracy of around 93%. The final accuracy of our proposed Threat Detection Model (TDM) is 82.65%.

**Keywords:** Accident Prediction Model, Autonomous Self Driving Car, Advanced Lane Detection, Computer Vision, COCO Dataset, Depth Estimation, Multi Regression, Two-Second Rule, YOLO.

## 1 INTRODUCTION

As technology is advancing very fast nowadays, therefore, whatever is relevant to the market today becomes irrelevant for tomorrow. Self-Driving Cars are the future and a very convenient and easy-to-use mode of transportation, however, the only reason that refrains people from adopting the same is its safety parameter. The concern includes if the car will be as safe as a traditional car and secondly, it can detect the threats accurately as well as alert the system proactively. To answer these questions, a substantial amount of research has been carried out in this field focusing on various approaches. Accidents have been on a surge in recent years and are constantly increasing over the years as the number of automobiles is increasing. The WHO in its recent report published in June 2021 [3], mentions some key facts which include that road accident deaths are around 1.4 million, GDP loss is in the range of 3% and also the most vulnerable age group is 5 to 29. Moreover, India has suffered a loss of between 1.17 lakh crores to 2.91 lakh crores in 2019 due to road accidents. India, which owns nearly 1% of global vehicles, contributes to around 11% of annual deaths in accidents [4]. Thus, having an automated system of vehicles that follow all the safety guidelines can be an effective solution to this problem as well.

The main objective of our work is to identify the threats and make the vehicle self-aware. The problem will be analyzed by taking into consideration various factors and varying the parameters such as *depth estimation* and *Region of Interest* (RoI) to achieve the best possible accuracy of the model. Adopting such a system will allow immediate response to threats using neural networks and computer vision. Moreover, we can directly use the model to deploy a system on cars that works in real-time and can detect accidents and provide indications that could help in reducing fatal injuries. The dataset has been sequentially analyzed, combined, and tested using the best-suited algorithm such as YOLO which uses the COCO dataset.

The paper is divided into four sections as follows:

1. Section 2 details an extensive literature survey of past studies.
2. Section 3 describes the flow of the proposed model and its implementation.
3. Section 4 presents the results and analysis of the proposed approach undervarying parameters.
4. Section 5 mentions the conclusion, limitations, and future scope.

## 2    BACKGROUND AND RELATED WORK

In this section, we present the literature review, summarize and analyze the work related to accident/threat detection in automated vehicles.

Deeksha Gour et al. [1] proposed an artificial intelligence-based traffic monitoring system that can detect the occurrence of accidents of vehicles in live camera feeds and detect collision of these moving objects. Further, the emergency alerts are immediately sent to the nearby authority for necessary actions. The paper focuses on the Yolo algorithm that is capable of detecting accidents and also it can run on CPU-based devices such as laptops or mobile phones. Here, the authors do not consider the actual distance of the vehicle from the camera which is an important parameter and can result in false signals being sent to vehicles moving in other lanes.

Ahmed M. Ibrahim et al. [2] presented the work which focuses on detecting the cars using YOLO and further assigning the cars into lanes using a real-time computer vision algorithm. For the proposed approach, the authors have divided the road into three sections, namely, left, right and emergency lanes which is the main region of focus to run the algorithm. This work suffers from limitations like the model could focus on car detection only and not other obstructions that may hinder the road. Also, image processing is done to consider only the car's lane and ignore all the obstructions that fall in the peripheral vision of the driver.

Nejdet Dogru et al. [5] proposed a system that detects the event of accidents by collecting necessary information like vehicle position, speed and other related abnormal activities from neighboring vehicles and process the same using machine learning tools to detect possible accidents in the future. The study aims to analyze traffic behavior and consider vehicles that move differently than current traffic behavior, as a possible accident case. The authors implemented a clustering algorithm to detect accidents, however, the problem is unsolvable with their approach when the road is curved as the data values will vary accordingly and this factor is not considered.

Vipul Rana et al. [6] presented a paper on accident detection and severity prediction with the help of machine learning and computer vision. The principal focus of the project is to detect the vehicles using the YOLO model and then perform watershed image segmentation using computer vision along with Random Forest Classifier. The project primarily focuses on alerting the drivers about the accident-prone areas and further reducing the chain of accidents rather than focusing on the driver's path.

Chaitali Khandbahale et al. [7] published a paper in which the authors used an embedded system within the vehicle. If the obstacle lies in the front, then the microcontroller can communicate inside the system and alert the driver, in case, any factor enters a non-safe zone. The main drawback is that sensors can be too costly and may not work correctly in rainy conditions.

Varsha Goud et al. [8] exhibited some research for accident detection using an accelerometer. With signals from an accelerometer, a severe accident can be recognized. According to this project report, whenever a vehicle meets with an accident, the vibration sensor will immediately detect the signal, or if a car rolls over, a micro-electro-mechanical system (MEMS) sensor will alert the system. The limitation comes in the form of costly sensors as well as the ,threat detection is possible only up to a certain distance.

T Kalyani et al. [9] proposed a model which can detect any accident on occurrence using vibrations and will send a message to a number. The hardware is used in the accident detection technique wherein at the time of an accident a vibration sensor is triggered due to vibration, the GPS module is then fetched and the location of the incident is sent as a message to the provided mobile number with the help of GSM module that is inbuilt into the particular system. The main problem here is that hardware is not always reliable and can cause a delay in predicting the nature of an accident.

Nikki Kattukkaran et al. [10] presented the idea of using an accelerometer. Here, the accelerometer in the vehicle senses the tilt of the vehicle and the heartbeat sensor on the user's body senses the abnormality of the heartbeat to understand the seriousness of the accident. Thus, the system made the decision and sent the information to the smartphone, through Bluetooth. However, the Bluetooth range is very limited and heartbeat is not a suitable measurement for accident verification because it is subjective to interpretation.

Xi Jianfeng et al. [11] published a work in which a classification and recognition method is used for analyzing the severity of road traffic accidents based on rough set theory and support vector machine (SVM). The authors classified the data based on attributes in humans, vehicles, roads, environment, and accidents. The main shortcoming is that SVM is found to be limited while analyzing objects and instance segmentation during detection and hence the instance classification is hindered.

M. Gluhakovi´c et al. [17] proposed a method for vehicle detection near an autonomous vehicle. The authors used YOLO v2 which employs identity mapping, concatenating feature maps from a previous layer to capture low-level features. As YOLO v2's architecture lacks in capturing a few important elements, we use Yolo v3 which uses residual blocks with no skip connections and no upsampling, thus, resulting in improvised modeling.

## 3   Methodology

This section proposes a four-step model for obstacle detection on the road using computer vision and neural networks. The camera mounted on the windshield records the video which is simultaneously executed on the deep learning algorithms. As an output from the proposed model, we successfully outline the obstacle with a red boundary and ensure to proactively alert the system about a potential threat on the road.

### 3.1   System Overview

The subsection briefly explains the model assumptions followed by broader steps of the proposed approach.

*Model Assumptions:*

1. The camera must be mounted on the windshield.
2. Focal length of the camera.
3. Speed must be provided while the vehicle is moving.

*Broader Steps:*
**Step 1: Region of Interest and Lane Detection**

The initial step is to identify the region of interest of the image by varying the parameters used for determining the horizon and also take some help from the lane detection algorithm.

**Step 2: Object Detection using YOLO**

Followed by identifying the main area to focus on, we will detect the obstacles in our region and also find their proximity from our region of interest so that we can anticipate any danger beforehand. In case, there is a real danger, it is outlined with a red boundary.

**Step 3: Distance of the obstacle from camera**

Next, we will calculate the distance of the object from our camera. Although there are many multi regression techniques to identify the same, we have used regression technique along with real-time data to calculate the model equation.

**Step 4: Calculating Safety**

For identifying the safety parameter, we used a popular two-second rule technique [16]. We will make use of the distance calculated in the previous step and the speed of our car to measure safety.

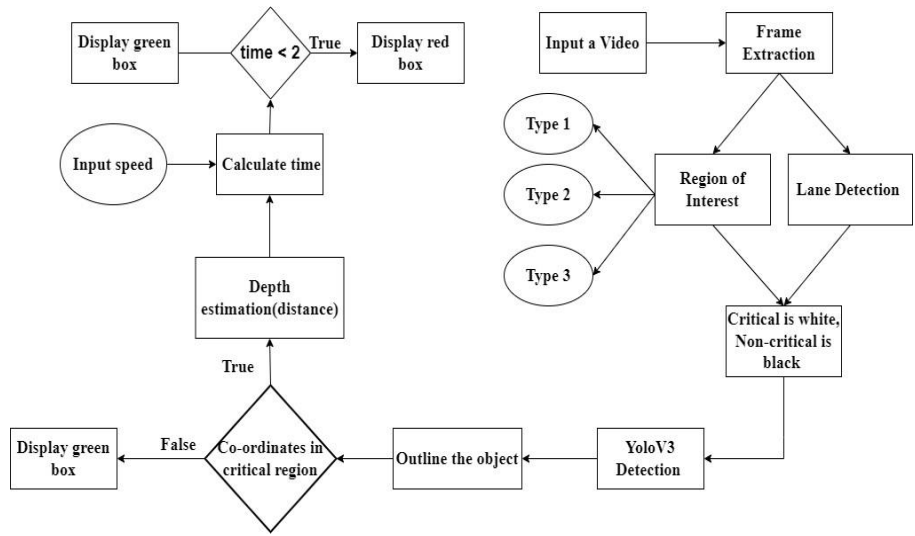

Fig.1: System Workflow

## 3.2    System Architecture

The subsection will discuss a detailed description of the steps of the proposed model. The workflow for the same has been shown in Figure 1.
**Lane Detection and Region of Interest**

In case, there are lines already drawn on the road, an advanced lane detection algorithm may be used. However, in the absence of lines, we may use an appropriate estimation method in order to calculate the horizon and then configure the lane.

   **i) Lane Detection:** Given an image as shown in Figure 2(a), the following steps are essential to determine the exact lane of the car:

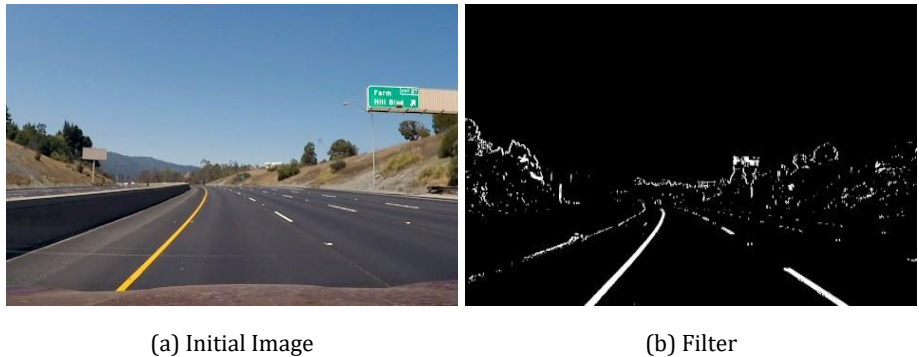

(a) Initial Image                              (b) Filter

Fig.2: Initial Image and Filter

Step 1: *Eliminating the distortions from the image:* The image is transformed into 2-dimensional objects to eliminate two sorts of distortions, i.e., radial and tangential We have used pythons' OpenCV library for the following purpose (Figure 2(b)).

Step 2: *Processing color channels:* Here, we obtain a binary image that contains the lane by applying filters on it (Figure 2(b)). For filtering, the three threshold criteria have been used; i) lightness threshold to remove the edges formed by shadows; ii) using saturation channel threshold to extract white lanes; iii) hue for the line colors.

Step 3: *Birds eye view:* Here, we wrap the image to have a top view of the lane. This helped us to effectively run our algorithm and fit a curve on lane pixels.

Step 4: *Finding left and right lane pixels:* We fit these pixels in a 2-degree polynomial. The value of the pixel is 0 and 1. We get peaks where the color is white and the rest of the time it is 0. From the peaks, the position of the lanes can be easily identified.

Step 5: *Sliding window:* From the base position of the left and right lines, we create a sliding window that is placed around the center of the line. This window finds and tracks the line to the top of the frame (Figure 3(a))

Step 6: *Determining the lane curvature:* A derivative function has been used to calculate the radius of curvature of the road [15].

Step 7: *Drawing the lane boundaries:* Finally, after executing the above steps, we will draw these white points on the white part which represents the lane on the original image (Figure 3(b)).

---

**Algorithm 1** Lane Detection

```
 1: #Calibrate and correct distortions
 2: undist = cv2.undistort(input image, mtx, dist, None, mtx)
 3: #process gradients and color channels for creation of binary image that consists of lane
      pixels.
 4: thresh img = ColorThreshCombined(undist, s thresh, l thresh, v thresh, b thresh)
 5: #Birds eye view
 6: warpped img = PerspectiveTransform(thresh img, src pts, dst pts, False)
 7: #process gradients and color channels for creation of binary image that consists of lane
      pixels.
 8: #Finding the lines
 9: left fit,right fit,lines img,mean curverad,position = find lines video(warpped img)
10: #unwarp the image back to the original perspective
11: inv matrix, unwarpped img = InvertPerspective(warpped img, src pts, dst pts)
```

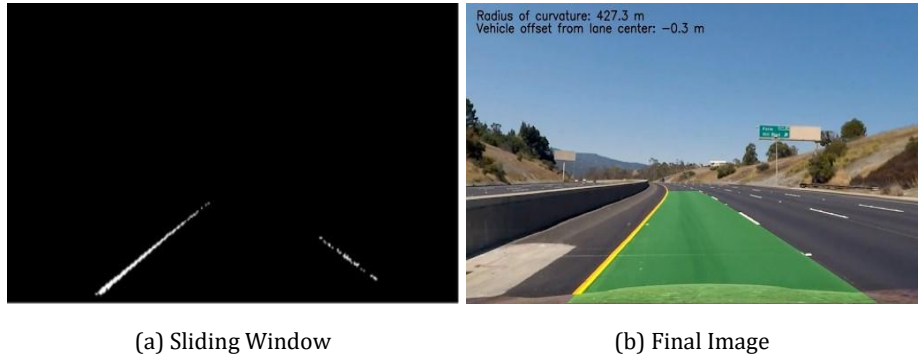

(a) Sliding Window          (b) Final Image

Fig.3: Sliding window and final Image

**ii) Region of Interest:** In case, there are no lanes on the road (which is common in many middle and small-income countries), our algorithm may not perform accordingly. In order to address the issue, we used geometric logic to identify the region of interest by varying the parameters. The fact is that a road in front of our eyes converges at infinity. Such a region is to be selected in a manner that the results must be accurate and can detect a threat quickly.

Using this logic, we divided the region into two parts initially, one is above and the other is below the horizontal line A as mentioned in Figure 4. The part which is above the line A in Figure 4. is not of major concern majority of the road lies in the region below line A. We are dividing the upper and lower part into a 45:55 split. If the center of the detected object is found in the upper part, it is rejected. Further, the parameters are varied to determine whether an object is a threat or not. Three such possibilities have been configured in Figure 4.

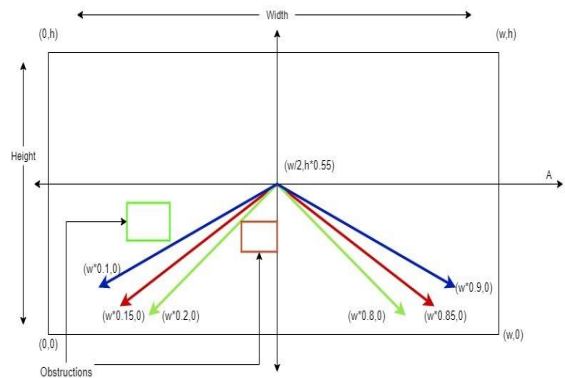

Fig.4: Region of Interest

1. The blue line in the figure above represents the first region of interest. It is the largest among all three and will detect most of the obstacles. The limitation is that some obstacles in certain situations may not be detected which do not pose a real danger.
2. The red line is a medium-sized region of interest and it may detect most of the objects. The final results are described in the next session.
3. The green line is the smallest of all and is least prone to detecting false danger; the only challenge is whether it will detect all the threats beforehand.

---

**Algorithm 2** Region of Interest

1: masked image = np.zeros like(InputImage)
2: cv2.fillPoly(masked image, RegionPoints, color =(255,255,255))

---

**Determining the obstacles using YOLO:**

YOLO [14] is the real-time object detection system that uses the COCO dataset from Darknet for training. This trained model will be able to detect vehicles from the roads in the working environment. Unlike classifier-based approaches, YOLO with the help of a loss function is trained and directly corresponds to detection performance and the whole model is trained together. YOLO outperforms the more accurate Fast RCNN as R-CNN takes about 2-3 seconds to predict an object in an image and hence is not feasible for real-time object detection. With the help of YOLO, we can train the whole object detection model with a single neural network.

YOLO when combined with several algorithms, such as lane detection, depth estimation, and computer vision can increase the likelihood to attain the desired threat detection or warning as mentioned in Figure 7. YOLO turns out to be ideal since it's robust, powerful, and quick object detection.

---

**Algorithm 3** YOLODetection

1: #configure Convolutional neural network using configuration and weights file with CPU related executions
2: neural network = cv2.dnn.readNetFromDarknet('yolov3.cfg', 'yolov3.weights')
3: neural network.setPreferableBackend(cv2.dnn.DNN BACKEND OPENCV)
4: neural network.setPreferableTarget(cv2.dnn.DNN TARGET CPU)

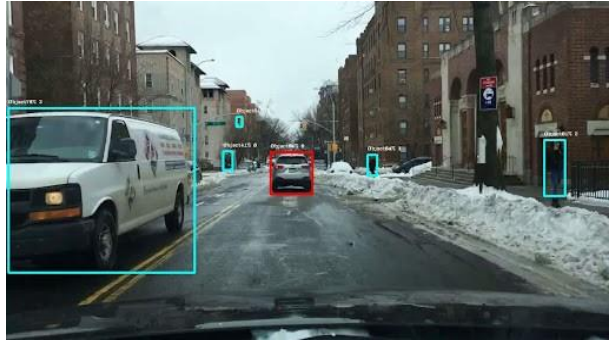

Fig.5: Yolo Detection Image

**Distance of the obstacle from camera**

In this section, we conducted an experiment to estimate the relationship and generate an equation between distance from the front camera of the car to any other obstruction detected in front of us in the danger region, where the selection of ROI/lane detection has been explained in the previous section.

Table 1: Variation of area as the distance changes

| Area (pixel$^2$) | Distance ($m$) |
|---|---|
| 440380 | 2 |
| 239598 | 3 |
| 137138 | 4 |
| 96657 | 5 |
| 67626 | 6 |
| 47294 | 8 |
| 33631 | 10 |
| 25479 | 11 |
| 12168 | 16 |

As you can see in the above Table 1, we choose mainly two parameters and conduct our experiment solely on these parameters.
i) Area(pixel$^2$): The area occupied in the image by obstruction (Refer figure 8 below). Here, the yellow box around the car rightly displays the area that is occupied by the car. ii) Distance($m$): The distance between the object and the front camera. In table 1, we placed a car at various distances from the car, captured a picture from that point, and took 9 different sets of photos. After capturing the image, our next step was to write a program that detects the car using YOLO and returns the coordinates of the car around which we not only draw a bounding box but also calculate the area and display it.

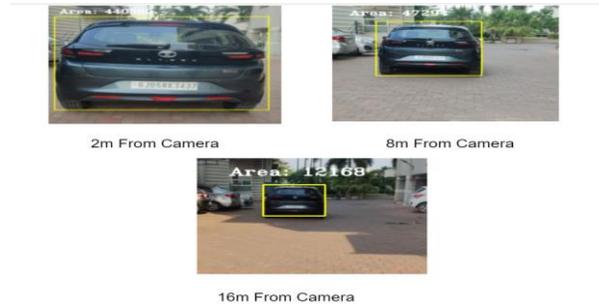

Fig.6: Depth Image

For readings in table 1, we parked a car at different distances from another car, captured a picture from one point and took 9 sets of photos. After capturing the image, we modeled our approach to detect the car using YOLO and returned the coordinates of the car around. Using this, we can draw a bounding box and calculate the area as well. Thereafter, a graph is plotted with the retrieved values wherein the x-axis depicts the area and y-axis depicts the distance. Finally, we chose a function which is the best fit for our observed values, i.e., the power function which generated an equation $Y=4319.3x^{-0.589}$. Afterwards, we integrate this in our system by just calculating the area of all the objects that are detected in the danger region, and further substituting that value in the generated equation which, in turn, gives the distance of the object from the car, which was the primary objective.

**Calculating the safety standards between two objects**

After calculating the objects in the region of interest and their distance from the camera, it is to identify whether the object is the real danger. For this, we follow the two-second rule, commonly used in many western and European countries. According to the rule, if a vehicle crosses point A, the vehicle following it must pass point A only after 2 seconds. Refer Figure 7.

---

**Algorithm 4** Calculating safety with the help of depth estimation and the 2-second rule

---

**if** Object Lies in critical region(CenterOfObject **then**
    depth   of   object=DepthEstimation(height,width)
    time = depth of object/input speed **if** Time < 2 **then**
    DANGER!!!
**else**
    SAFE!

## 4   OBSERVATIONS AND RESULTS

In this work, we consider two important parameters to be varied for accuracy, namely, *region of interest* and *depth estimation*. In order to calculate the accuracy, we downloaded the dataset [12] and used 500 videos of a car being driven, wherein 250 videos showed the car crash and the remaining 250 videos showed no car crash. The length of each video is 10-15 seconds on an average. Further, each video is fetched as an input in our system, and later on checked if the system correctly displayed the red bounding box/ blue bounding box, i.e., the two classification of output which is to ensure if a vehicle is under a threat. The classification of Region of Interest (ROI) in our work is as:

### 4.1   Region of Interest

i)  *Type 1* : It is the biggest of all the 3 types having the base coordinates at
0.1*w and 0.9*w, respectively, (Blue line).
ii) *Type 2* : It is medium in size and has base coordinates at a distance 0.7*w
   apart at 0.15*w and 0.85*w, respectively, (Red line). iii) *Type 3* : The smallest
   of all and has its base coordinates at 0.2*w and
0.8*w, respectively, (Green line).

**Three ROI's and their accuracy**

    In Table 2, the accuracy in Type 1 ROI is comparatively less as the angle is more, therefore, it unnecessarily detects and run the algorithm on outliers. The accuracy of Type 3 is less than the accuracy of Type 2 mainly because the angle is very narrow and hence it is unable to detect the cars that are in the peripheral vision of the camera, thereby, eliminating objects in vicinity of the car. Therefore, we can conclude that the accuracy is best amongst the three in Type 2.
    The reason behind the calculations of this type of results was mainly to understand how varying angles in ROI could play a significant role in establishing the system.

Table 2: Comparison of accuracy in different Region of Interest

| Region of interest | Accuracy (in percentage) |
|---|---|
| Type 1 | 71.42% |
| Type 2 | 82.65% |
| Type 3 | 79.59% |

**Three ROI's and their frames**

The analysis in Table 3 can help us to understand how quickly an output will be generated and display the results to the user. Here, the frame delay is calculated in the output of the video and the results are obtained by providing 250 video inputs which fall under category 1, i.e., the car crashes. Thereafter, we find the number of frames until the system is able to find the object and generate a result for the first time for all 250 videos and calculate the average and median for the same.

Type 1 is having the largest angle and it would naturally generate results faster as it will detect the objects quicker. Similarly, the angle becomes shorter in Table 3: Comparison of Average and Median in Different Region of Interest

| Region of interest | Frame Delay (Average) | Frame Delay (Median) |
|---|---|---|
| Type 1 | 10.71 | 7 |
| Type 2 | 11.41 | 8 |
| Type 3 | 12.35 | 10 |

Type 2, thereby, resulting in more average frame delay as compared to Type 1. The frame delay is comparatively the most in Type 3 as the angle is the shortest, therefore, it takes more time than other techniques to actually detect an object and generate results. The comparison can be made in only those videos where all the 3 regions of interest give accurate results, so the test dataset videos were reduced to around 250. The average frames per video is 38.

## 4.2   Depth estimation

As mentioned briefly in section 3, we carried our own experiment and noted the observations. Afterwards, we plotted the data on a scatter plot and finally made two best fit equations. Firstly, by using a 2-degree polynomial expression followed by an exponential expression. Refer Table 4.

Here, we found out that the polynomial expression gave a less accurate result (75.79%) as compared to exponential (82.65%) as shown in Figure 8(a) and Figure 8(b), respectively. One of the reasons might be that in a certain range (250000-400000), the polynomial expression exhibits a very less distance and resulted in detecting some vehicles which were not a problem or threat. Also, the expression indicates a big distance for a very large area which can also be a possibility of less accuracy to some extent.

Table 4: Comparison of Accuracy Using Different Equations

| Equation | Accuracy using ROI 2 |
|---|---|
| Polynomial (y=1E-10$x^2$-9E-05x+13.721) | 75.61% |
| Exponential (y=4319.3$x^{-0.589}$) | 82.65% |

## 5  Conclusion and Future Scope

Automated cars are the need of the hour. However, the main hindrance is reliability and safety. With this research work, we aimed to focus on obstacle detection techniques. Our proposed system was divided into four sections, i.e., finding the lane/ROI, detecting the cars, calculating the distance using various depth estimation techniques and calculating safety. During this process, ROI having coordinates 0.7w apart at 0.15w and 0.85w along with an exponential equation for depth estimation resulted in positive threat detection with 82.65% accuracy. For future work, we will focus on improvement of the accuracy as well as ensure that the model learns and reacts to more complex situations like threats from behind and sideways. Also, some complex cases of curvature can be considered for reduced delay in threat identification and notification.

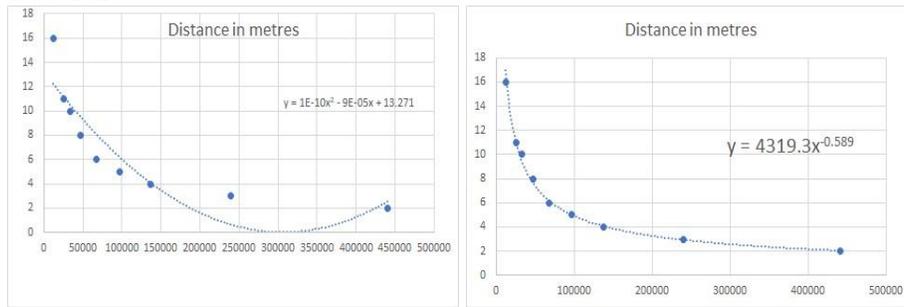

(a) Polynomial Plot           (b) Exponential Plot

Fig.7: Graph of Depth estimation equations